\documentclass[conference]{IEEEtran}
\IEEEoverridecommandlockouts

\usepackage{cite}
\usepackage{amsmath,amssymb,amsfonts}
\usepackage{graphicx}
\usepackage{textcomp}
\usepackage{xcolor}
\usepackage[symbol]{footmisc}
\def\BibTeX{{\rm B\kern-.05em{\sc i\kern-.025em b}\kern-.08em
    T\kern-.1667em\lower.7ex\hbox{E}\kern-.125emX}}

\usepackage{algorithm}
\usepackage{algorithmicx}
\usepackage{algpseudocode}
\usepackage{booktabs}

\usepackage{amsmath}
\usepackage{multirow}
\usepackage{color}
\usepackage{makecell}
\usepackage{xcolor}
\usepackage{arydshln} 
\usepackage{tabularx}
\usepackage{subcaption}
\usepackage{hyperref} 
\hypersetup{
    colorlinks=false,     
    pdfborder={0 0 0},    
}
\newcommand{\smnormalsize}{\fontsize{17pt}{11pt}\selectfont}
\newcommand{\ustitle}{T{\smnormalsize AB}F{\smnormalsize AIR}GDT}
\newcommand{\us}{T{\scriptsize AB}F{\scriptsize AIR}GDT}
\newcommand{\mostlyai}{TabularARGN}
\newcommand{\tabfairgan}{TabFairGAN}
\newcommand{\cuts}{CuTS}
\newcommand{\fsmote}{FSMOTE}
\newcommand{\prefair}{PreFair}

\begin{document}

\title{\ustitle: A Fast Fair Tabular Data Generator using Autoregressive Decision Trees\\
{}
}


\author{
\IEEEauthorblockN{
Emmanouil Panagiotou\IEEEauthorrefmark{2}\IEEEauthorrefmark{4}\IEEEauthorrefmark{1},
Beno\^it Ronval\IEEEauthorrefmark{3}\IEEEauthorrefmark{1},
Arjun Roy\IEEEauthorrefmark{2}\IEEEauthorrefmark{4},\\
Ludwig Bothmann\IEEEauthorrefmark{5},
Bernd Bischl\IEEEauthorrefmark{5},
Siegfried Nijssen\IEEEauthorrefmark{3}\IEEEauthorrefmark{6},
and Eirini Ntoutsi\IEEEauthorrefmark{4}}

\IEEEauthorblockA{\IEEEauthorrefmark{2}
Freie Universität Berlin,
Berlin, Germany\\
}

\IEEEauthorblockA{\IEEEauthorrefmark{3}
ICTEAM, UCLouvain,
Louvain-la-Neuve, Belgium\\
}

\IEEEauthorblockA{\IEEEauthorrefmark{4}
Research Institute CODE, Universität der Bundeswehr München, Munich, Germany
}

\IEEEauthorblockA{\IEEEauthorrefmark{5}
Department of Statistics, Munich Center for Machine Learning (MCML),
LMU Munich, Munich, Germany}

\IEEEauthorblockA{\IEEEauthorrefmark{6}
DTAI,
KU Leuven, Leuven, Belgium}

\IEEEauthorblockA{\IEEEauthorrefmark{1}Authors contributed equally to this research}

\IEEEauthorblockA{\IEEEauthorrefmark{0}
Correspondence: emmanouil.panagiotou@fu-berlin.de, benoit.ronval@uclouvain.be}

}

\maketitle

\begin{abstract}

Ensuring fairness in machine learning remains a significant challenge, as models often inherit biases from their training data. 
Generative models have recently emerged as a promising approach to mitigate bias at the data level while preserving utility. However, many rely on deep architectures, despite evidence that simpler models can be highly effective for tabular data.
In this work, we introduce~\us, a novel method for generating fair synthetic tabular data using autoregressive decision trees. To enforce fairness, we propose a soft leaf resampling technique that adjusts decision tree outputs to reduce bias while preserving predictive performance. 
Our approach is non-parametric, effectively capturing complex relationships between mixed feature types, without relying on assumptions about the underlying data distributions. We evaluate \us\ on benchmark fairness datasets and demonstrate that it outperforms state-of-the-art (SOTA) deep generative models, achieving better fairness-utility trade-off for downstream tasks, as well as higher synthetic data quality. Moreover, our method is lightweight, highly efficient, and CPU-compatible, requiring no data pre-processing. Remarkably, \us\ achieves a 72\% average speedup over the fastest SOTA baseline across various dataset sizes, and can generate fair synthetic data for medium-sized datasets (10 features, 10K samples) in just one second on a standard CPU, making it an ideal solution for real-world fairness-sensitive applications.


\end{abstract}

\section{Introduction}
Machine learning (ML) models are increasingly deployed in critical sectors like finance, healthcare, and education. 
However, these models often showcase bias towards individuals or specific demographic groups based on sensitive attributes such as race, gender, age, or socio-economic status.
Such biases can arise from various sources \cite{mehrabi_survey_2022}, including historical discrimination, data collection methodologies \cite{schwartz2022towards}, and \textit{algorithmic biases}, where design choices within the algorithm, such as the optimization function, introduce bias despite unbiased input data. One potential discriminatory outcome is \emph{group (un-) fairness}, where certain demographic groups receive biased predictions. This occurs when the model systematically favors or disadvantages certain groups based on sensitive attributes. As a result, ensuring fairness in ML models is essential, and a substantial body of literature has been devoted to addressing this issue, ranging from basic philosophical frameworks \cite{bothmann_what_2024} to technical solutions, typically categorized into pre-, in-, and, post-processing approaches, proposing interventions for fairness at the data-, algorithm-, or model-level, respectively  \cite{caton_fairness_2023}. 
We focus on approaches that address fairness at the data level, particularly for tabular data, which is the most prevalent data type in real-world applications \cite{chui2018notes}.

Fairness at the data level involves modifying the training data ``towards fairness'' before feeding them to an ML algorithm. The notion of fair data is defined in various ways in the literature, depending on the specific application or context. Most often, it is framed as achieving statistical parity with respect to a known sensitive attribute and a specific target variable \cite{rajabi2022tabfairgan,krchova2023strong}. However, alternative definitions exist, such as, counterfactual fairness \cite{leininger2025overcoming}, $\epsilon$-fairness, when fairness is pursued without a specific downstream target \cite{feldman2015certifying}, or even in situations where sensitive attributes are not observed \cite{bothmann_what_2024}. To achieve fair data, various pre-processing approaches have been used, by transforming, reweighting, massaging class labels~\cite{tawakuli2024make}, or augmenting data to mitigate bias, ensuring that sensitive attributes do not unfairly influence predictions. While traditional augmentation techniques, such as oversampling~\cite{SONODA2023119059,ronval2024can,panagiotou2024synthetic} or reweighting, adjust the existing dataset, generative models~\cite{rajabi2022tabfairgan,tiwald2025tabularargn} offer a more advanced alternative by synthesizing entirely new data that adheres to fairness constraints, preserving statistical properties while reducing bias. Most such methods rely on deep architectures, often leading to longer computational times, even when using GPUs. However, simpler models have proven to remain effective for tabular data generation, while requiring fewer resources in terms of both time and computational power \cite{panagiotou2024synthetic,ronval2024can}.

In this work, we focus on \emph{generating} fair synthetic data, following a common assumption in synthetic data generation for fairness~\cite{rajabi2022tabfairgan,krchova2023strong,SONODA2023119059,vero2023cuts}, where both the sensitive attribute and the target attribute are observed. The goal is to generate fair data that (i) preserves the statistical properties of the original data (ii) eliminates discrimination, and (iii) maintains utility on the downstream task. To achieve this, we propose  \us\ (\emph{\underline{Tab}ular \underline{Fair} \underline{G}enerative \underline{D}ecision \underline{T}rees}). Our approach follows an autoregressive framework, generating attributes sequentially using decision trees (DTs). To ensure fairness, we incorporate a \emph{fair leaf resampling} algorithm in the final step when generating the target attribute. This intervention enforces statistical independence between the target and the sensitive attribute in the generated data while preserving the underlying attribute distributions. \us\ is a non-parametric approach, making no assumptions about the distributions of attributes, and needs no pre-processing, making it ideal for mixed tabular data, which are common in this domain~\cite{le2022survey}. Our method is simple yet highly effective, consistently maintaining utility, measured by downstream ROC AUC performance, and data quality, while significantly improving fairness, assessed by downstream statistical parity. 
Experimental evaluation on fairness benchmark datasets confirms that \us\ achieves superior results compared to existing approaches. Furthermore, it is efficient due to the use of DTs, generating fair synthetic data for medium-sized datasets in just one second on a standard computer. Our approach is also lightweight and CPU-compatible, eliminating the need for specialized hardware. All these advantages make \us\ ideal for real-world applications. 
We provide a codebase\footnote{\us\ code: \href{https://github.com/Panagiotou/TABFAIRGDT}{github.com/Panagiotou/TABFAIRGDT}} to reproduce all experiments.


In summary, the contributions of this work are the following:
\begin{itemize}
    \item We introduce \textbf{\us}, a novel fair synthetic data generation method using an autoregressive decision tree (DT) framework. It is \textit{efficient} and \textit{non-parametric}, requiring no pre-processing or specialized hardware. 
    \item We propose a \textit{fair leaf resampling} algorithm to enforce fairness while preserving data quality and utility.  
    \item \us\ outperforms existing approaches in \emph{fairness-utility tradeoff}, improving fairness by $\raisebox{0.3ex}{\scriptsize$\sim$}\, 50\%$ on average while preserving high predictive score.
    \item Our method avoids generating sensitive attribute-specific out-of-distribution samples, making it suitable for real-world critical applications like healthcare.  
\end{itemize}

\section{Related work}


\label{sec:related_work}

Recent years have seen an increase in contributions tackling fairness issues in machine learning. Metrics to measure (un-)fairness can be divided \cite{verma_fairness_2018} into group and individual fairness. Methods to address (un-)fairness, on the other hand, can be divided into pre-processing, in-processing, and post-processing \cite{caton_fairness_2023}. Our method falls into the category of pre-processing methods, generating fair synthetic data that subsequently can be used for creating models adhering to group fairness. Although there are multiple ways to evaluate fairness, we follow related work for fair synthetic data generation, primarily reporting on statistical parity \cite{rajabi2022tabfairgan,krchova2023strong,SONODA2023119059,vero2023cuts}, measuring the difference in positive outcome rate between the sensitive attribute groups (c.f. Section \ref{sec:metrics} for a formal definition). Hereafter, we analyze works relevant to our method, focusing on generative models for tabular data and their fairness-aware variations. Furthermore, since our method is tree-based, we provide an overview of fair DTs.


\noindent\textbf{Generative Models for Tabular Data:}
Generative models have been used for different data types, from images to text and audio, including tabular data. One family of methods adapts deep learning architectures from other domains for tabular data. For example, CTGAN and TVAE \cite{xu2019modeling} use Generative Adversarial Networks (GANs) and Variational Auto-Encoders to generate tabular data. More recently, diffusion-based models have outperformed them by reversing a gradual noising process to reconstruct tabular data, either directly in input space \cite{kotelnikov2023tabddpm} or via latent diffusion \cite{zhang2023mixed}. Other categories of methods exist that do not specifically rely on deep learning architectures. For instance, by approximating the joint distribution of tabular data using $n$-way marginals \cite{mckenna2022aim}, estimating densities through random forests \cite{watson2023adversarial}, or employing distance-based interpolation \cite{chawla2002smote}. In \cite{reiter2005using}, a generative model utilizing Classification and Regression Trees (CART) is introduced for sequential column-wise data generation. Our method builds upon this approach by integrating fairness constraints.

\noindent\textbf{Fair Generative Models for Tabular Data:}
Methods that introduce fair generative models are closely related to our work. Most approaches follow a two-step process: first, training a deep generative model, such as \tabfairgan\ \cite{rajabi2022tabfairgan} and \cuts\ \cite{vero2023cuts}, and then fine-tuning it for fairness using a regularized loss function. Similarly, \mostlyai\ \cite{krchova2023strong,tiwald2025tabularargn}, commercially developed by Mostly AI\footnote{\url{https://mostly.ai/}}, uses a deep autoregressive model and optimizes for fairness only when generating the target variable.
\prefair\ \cite{pujol2022prefair} utilizes marginals and a causal model to achieve fairness. Finally, some methods focus on fair data augmentation for the underrepresented groups \cite{panagiotou2024synthetic,ronval2024can,yan2020fair}.
Most of these approaches build upon SMOTE \cite{chawla2002smote}, such as \fsmote\ \cite{SONODA2023119059}. 
These fair data generators serve as baselines in our evaluation, and are described in detail in Section~\ref{sec:baseline_competitors}.

\noindent\textbf{Fair Decision Trees:} Our work closely relates to fairness-aware DTs, which incorporate fairness constraints into the DT induction process. Seminal work by~\cite{kamiran2010discrimination} introduced two strategies: (i) fair splitting criteria, which modify the tree construction process to enforce fairness during split selection, and (ii) leaf relabeling, which adjusts leaf labels post-construction to enhance fairness. Their study concluded that leaf relabeling was the most effective of the two strategies. Subsequently, \cite{10.5555/3367032.3367242} integrated a fair splitting criterion within Hoeffding trees to ensure fairness in streaming data. More recently, \cite{jovanovic2023fare} used a fair splitting criterion to facilitate fairness-aware representation learning. Our approach draws inspiration specifically from the leaf-relabeling strategy and employs a fair leaf-relabeling method to generate fair synthetic data. Although we also experimented with the fair splitting criterion, we found that it introduced substantial structural changes to the DT, compromising data utility.


\section{\us: Fast Fair Tabular Data Generation using Autoregressive Decision Trees}

\begin{figure*}[t]
    \centering
    \includegraphics[width=0.7\textwidth]{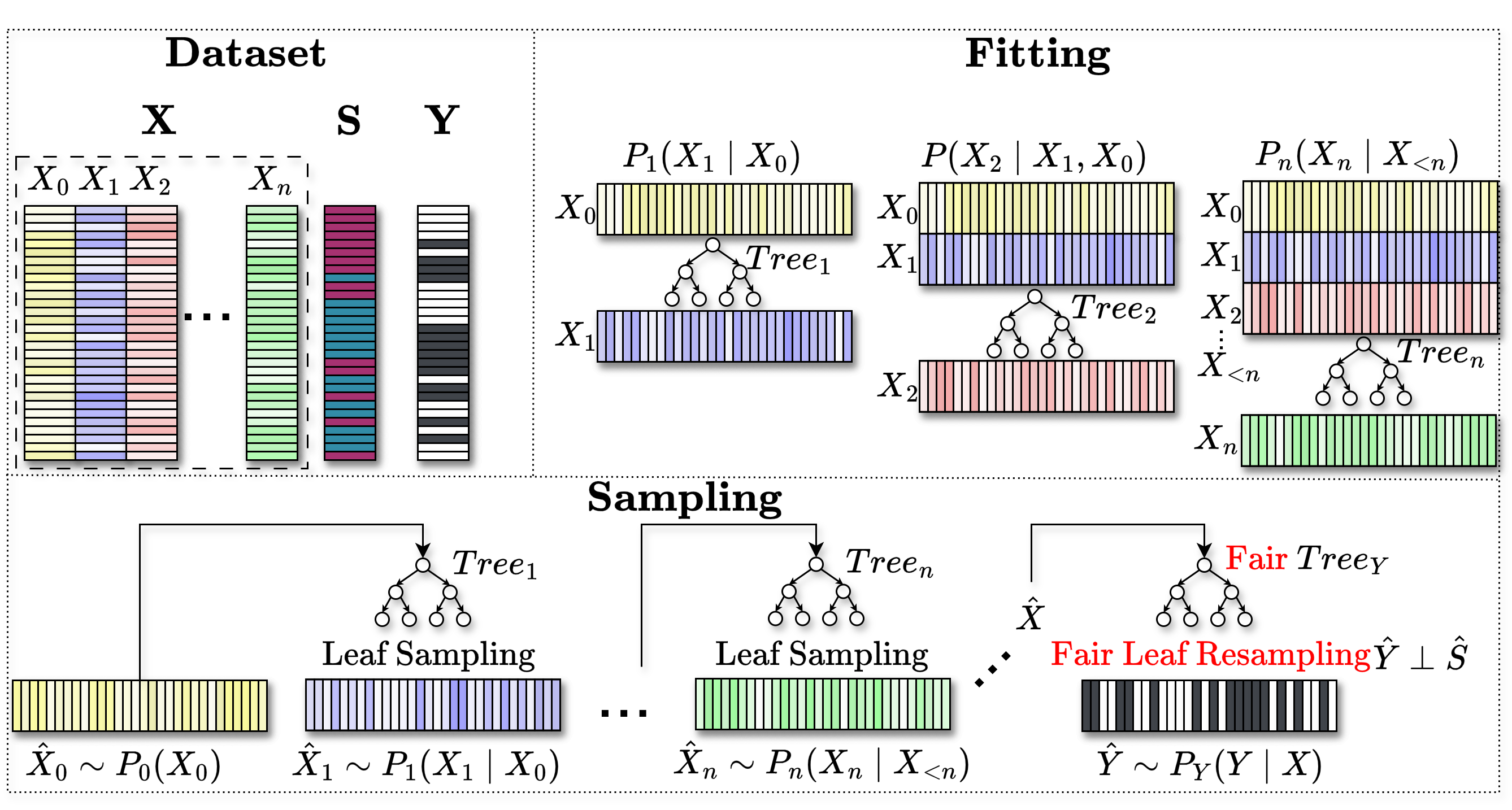}
    \caption{Overview of the synthetic data generation in \us.}
    \label{fig:us}
\end{figure*}

An overview of our approach is shown in Figure~\ref{fig:us}. 
Our method consists of two main components: an autoregressive tree-based generation step (Section~\ref{sec:architecture}), and a fairness-utility tradeoff-aware generation step (Section~\ref{sec:fair_generation}). 
Before presenting the components in detail, we first present the problem formulation in Section~\ref{sec:problem_def}.

\subsection{Problem Formulation}
\label{sec:problem_def}
\setcounter{footnote}{0}
\renewcommand{\thefootnote}{\fnsymbol{footnote}}
Let $D$ be a dataset defined as $D \subseteq X \times S \times Y$ where $X= X_0 \times X_1 \times \dots \times X_{n}$ represents 
$n$ numerical or categorical features, $S$ is a sensitive attribute, and $Y$ is the target attribute. 
Following common practice in fairness-aware learning methods and datasets~\cite{le2022survey}, we assume that both  $S\in\{0,1\}$ and $Y\in\{0,1\}$ are binary variables. For example, $S$ could represent $sex\in\{male, female\}$\footnote{Sex is considered binary in the datasets, not as a reflection of broader societal views.}, and $Y$ could represent $income\in\{low, high\}$. 

The goal of fair generative AI, as defined in the relevant literature \cite{rajabi2022tabfairgan,krchova2023strong,vero2023cuts}, is to learn a generative model based on $D$, capable of generating a synthetic dataset $\hat{D} \subseteq \hat{X} \times \hat{S} \times \hat{Y}\ $ that satisfies the following desiderata/requirements:
\begin{itemize}
    \item[\textbf{R1}] \textbf{- Data quality:} the generated data should approximate the original data, i.e., $\hat{X} \sim X$ and $\hat{S} \sim S$. 
    \item[\textbf{R2}] \textbf{- Fairness control:} the target should be statistically independent of the sensitive attribute, i.e., $\hat{Y} \perp \hat{S}$.
    \item[\textbf{R3}] \textbf{- Utility preservation:} the utility of the data for the downstream task should be preserved, i.e., $\hat{Y} \sim P(Y|X)$.
\end{itemize}


To achieve R1 (data quality), features are generated in an autoregressive fashion, where each feature is conditioned on the previously generated features (Section~\ref{sec:architecture}). R2 (fairness control) and R3 (utility preservation) are often in conflict~\cite{menon2018cost}, as statistical dependencies frequently exist between the sensitive attribute $S$ and the target variable $Y$ in the original dataset. Enforcing fairness (R2) by reducing or removing these dependencies can limit the model’s ability to preserve predictive performance (R3), thereby leading to an inherent trade-off between fairness and utility. We adress this trade-off through a fairness-aware target generation (Section~\ref{sec:ourFairRelabeling}).




\renewcommand{\thefootnote}{\arabic{footnote}}
\setcounter{footnote}{1}

\subsection{Autoregressive Tree-based Generation}
\label{sec:architecture}


The autoregressive synthetic data generation process can be formulated as sampling from a sequence of estimated conditional distributions, where each variable's synthetic value depends on the previously generated synthetic values. Given the dataset features $X_0, X_1, \cdots, X_n$, we  formally define the process as follows (\!\cite{uria2016neural,tiwald2025tabularargn,reiter2005using}):
\[
\begin{aligned}
\hat{X}_0 &\sim P_0(X_0) \\
\hat{X}_1 &\sim P_1(X_1 \mid X_0) \\
\hat{X}_2 &\sim P_2(X_2 \mid X_1, X_0) \\
&\;\;\vdots \\
\hat{X}_n &\sim P_n(X_n \mid X_{<n})
\end{aligned}
\]

\begin{algorithm}[t]
\caption{Decision Tree-based Synthetic Data Generation}
\label{algo:generation}
\begin{algorithmic}[1]
\Statex \textbf{Input} $X_{<j}$, real values for previous features; $X_{j}$ real values for current feature $j$; $\hat{X}_{<j}$ synthetic values for previous features. 
\vspace{0.2cm} 
\Statex \textbf{Output} $\hat{X}_{j}$ synthetic values for feature $j$. 
\vspace{-0.2cm} 
\Statex \hrulefill  
\Statex \textbf{i) Fitting:}
\State Train decision tree $Tree(X_{<j}) \rightarrow X_j$
\State Initialize probability dictionary $\mathcal{P} \gets \{\}$  
\For{each leaf $\ell$ in $Tree$}
    \State $p_{\ell}(X_j| X_{<j} \in \ell) := \frac{\text{count}(X_j = x_j \in \ell)}{\text{count}(X_j \in \ell)}\ \forall\ x_j \in X_j$ in $\ell$
    \State $\mathcal{P}[\ell] \gets p_{\ell}(X_j| X_{<j} \in \ell)$ \Comment{Store out probs per leaf}
\EndFor
\vspace{-0.2cm} 
\Statex \hrulefill  
\Statex {\textbf{ii) Sampling:}} 
\State Initialize dictionary $\mathcal{ID} \gets \{ \ell \to [\ ] \,|\, \ell \in Tree\}$  
\For{each index $i$ in $\{1, 2, \dots, |\hat{X}_{<j}|\}$} \Comment{Loop over all samples}
    \State $\hat{x} \gets \hat{X}_{<j}[i]$  
    \State Find leaf $\ell = Tree(\hat{x})$
    \State $\mathcal{ID}[\ell] \gets \mathcal{ID}[\ell] \cup \{i\}$  \Comment{Store indices of $\hat{X}_{<j}$ per leaf} 
\EndFor

\State $\hat{X}_j \gets [\ ]$

\For{each leaf $\ell$ in $\mathcal{ID}$} 
    \State $idxs \gets \mathcal{ID}[\ell]$ \Comment{Get all indices of $\hat{X}_{<j}$ in each leaf}
    \State $n \gets \text{length}(idxs)$  
    \State $\hat{X}_j[idxs] \gets \text{sample}(\mathcal{P}[\ell], n)$  \Comment{Draw $n$ synthetic samples from $p_{\ell}$} 
\EndFor

\State \textbf{Return} $\hat{X}_j$
\end{algorithmic}
\end{algorithm}

The autoregressive process generates each feature $X_j, j \in \{0, \dots, n\} $ sequentially, conditioned on prior synthetic values. Similarly, the sensitive attribute $S$ is generated, given all other columns $X$, i.e. $\hat{S} \sim P_S(S \mid X)$. 
To model the conditional probabilities $P_j$ for a feature $X_j$ we use DTs due to their efficiency, flexibility, and ability to handle diverse data types (\!\cite{reiter2005using}). The iterative per-feature generation process consists of two phases: fitting and sampling (c.f. Algorithm~\ref{algo:generation}):
\begin{itemize}
    \item[\textbf{(i)}] \textbf{Fitting phase}: For a feature $X_j$, a DT is trained to map $X_{<j}$ to $X_j$, storing the probability output distribution for each leaf in a dictionary $\mathcal{P}$, effectively estimating $P_j(X_j \mid X_{<j})$ (c.f. Algorithm~\ref{algo:generation} line 5).
    \item[\textbf{(ii)}] \textbf{Sampling phase}: Synthetic data of the previous columns $\hat{X}_{<j}$ is passed through the tree, and samples are drawn from $\mathcal{P}$ based on leaf assignments, to generate the next synthetic column $\hat{X}_{j}$ (c.f. Algorithm~\ref{algo:generation} line 17).
    \end{itemize}

The same fitting-sampling approach is used to generate all columns $\hat{X_j}$, except for the first feature, $\hat{X}_0$, which is sampled at random from $X_0$, since there are no previous synthetic values to generate from. This autoregressive generation process ensures the quality of data generation (R1).

\noindent \textbf{Discussion on the autoregressive generation process:} \us\ employs DTs at its core due to their simplicity, efficiency, and non-parametric nature, i.e. making no assumption on the underlying data distribution (as previously explored in  \cite{reiter2005using}).
In the context of fair data generation, simpler models have been shown to outperform more complex, generative approaches \cite{panagiotou2024synthetic,ronval2024can}.
Our generation process proceeds in the original order of features as present in the dataset. As shown in Section \ref{sec:ordering}, the feature generation order does \emph{not} significantly impact the results. Our autoregressive approach assumes correlations between certain features, as observed in real datasets. If all features are fully independent, the applicability of our approach would be limited, although such datasets are rarely useful in practice.

\subsection{Fairness-Aware Target Generation}
\label{sec:ourFairRelabeling}

\label{sec:fair_generation}
In the previous section, we described the autoregressive generation process using DTs, highlighting its flexibility, which allows to generate the sensitive attribute $\hat{S}$ after all other attributes have been produced. 
Moreover, it allows enforcing fairness constraints exclusively at the final step, when generating the target variable, ensuring fairness control (R2) and utility preservation (R3). 
Enforcing fairness earlier in the generation process may lead to out-of-distribution samples as observed in our experiments (c.f. Section~\ref{sec:female_husbands}). Such deviations directly reduce data quality (R1) and can be particularly problematic in high-stakes domains such as healthcare, where data fidelity is critical. Imposing fairness constraints only in the final step has also been supported by recent literature, including \cite{krchova2023strong,tiwald2025tabularargn}, which apply such constraints exclusively during target generation in deep autoregressive frameworks.


Our goal is to generate fair data by ensuring statistical parity in downstream tasks.
In this work, we focus on statistical parity as our fairness criterion, acknowledging that fairness is inherently domain-specific and that the appropriate fairness measure should be selected accordingly\footnote{In future work, we plan to incorporate alternative fairness notions.}.
Specifically, we generate the synthetic target variable $\hat{Y}$ so that a model's predictions $Y_{pred}$, satisfy statistical parity: $$P(Y_{pred}=1 | S=0) = P(Y_{pred}=1 | S=1)$$ To achieve this, we build upon the fair leaf relabeling approach~\cite{kamiran2010discrimination}, a post-processing bias mitigation technique that modifies a DT by performing hard relabeling of its leaf nodes to reduce discrimination. We extend this idea by introducing a novel soft resampling procedure that probabilistically samples $\hat{Y}$ based on a fairness-aware adjustment of the leaf predictions. Unlike hard relabeling, our approach, explained hereafter, enables smoother control over the fairness-utility trade-off. 



\noindent{\textbf{Fair leaf resampling:}} Using the final tree fitted on $X$ to predict $Y$ ($Tree_Y(X)$), we obtain a set of predicted labels $Y_l$. The discrimination and accuracy of the tree are defined as, $\text{disc}:=P(Y_{l}=1|S=0)-P(Y_{l}=1|S=1)$ and 
$\text{acc}:=P(Y_{l}=Y)$. For each leaf node $\ell$ in the tree, 
which currently predicts $y_{l} \in \{0,1\}$, we can estimate the potential impact on the total discrimination and accuracy of the tree, if the leaf's label were flipped to $y_{l}' = 1 - y_{l}$ as: $\Delta \text{disc}_\ell := P(y_{l}'=1|S=0, X \in \ell)-P(y_{l}'=1|S=1, X \in \ell)$, and $\Delta \text{acc}_\ell := P(y_{l}' \neq Y, X \in \ell)$.

Thus, the task of producing a fair tree can be formulated as an optimization problem: find a subset of leaf nodes $\mathcal{L}$ whose relabeling, under a discrimination threshold $thr_{disc}=0$, minimizes the overall discrimination (fairness control - R2) while incurring a minimal loss in accuracy (utility preservation - R3). Formally, the fair resampling objective is:
\begin{align*}
& \min_{\mathcal{L}} \left( -\sum_{\ell \in \mathcal{L}} \Delta \text{acc}_\ell \right)\\
\text{with new\_disc}(\mathcal{L}) &:= \text{disc} + \sum_{\ell \in \mathcal{L}} \Delta \text{disc}_\ell \leq \text{thr}_{\text{disc}}
\end{align*}


As proven in \cite{kamiran2010discrimination}, finding $\mathcal{L}$ can be reduced to a \emph{KNAPSACK} problem, allowing the use of a greedy algorithm (Algorithm~\ref{algo:resampling}-i), based on the discrimination-to-accuracy ratio of each leaf. However, when generating synthetic data from the leaves, we do not use the leaf's labels $y_{l}$; instead, we sample from the output probability distribution $p_{\ell}$ (Algorithm~\ref{algo:generation} line 17). 
Therefore, for each leaf in $\mathcal{L}$, instead of performing a hard relabeling, we compute adjusted sampling probabilities that reflect a trade-off between fairness and utility (Algorithm~\ref{algo:resampling}-ii). This trade-off is governed by a user-defined parameter $\lambda \in [0,1]$, commonly used in fair data generation. 
For example, consider a selected leaf $\ell \in \mathcal{L}$) with original output probabilities: $p_{\ell}=\{y_{l}=1:0.8, y_{l}=0:0.2\}$. For a complete switch ($\lambda=1.0$), the new probabilities will be: $p_{\ell}' =\{y_{l}=1:0.0, y_{l}=0:1.0\}$. For a softer adjustment ($\lambda=0.3$), the new probabilities will be: $p_{\ell}' =\{y_{l}=1:0.62, y_{l}=0:0.38\}$. 
Adjusted probabilities are then used to generate the target $\hat{Y}$.

To conclude, our fairness-aware target generation method selectively resamples a greedily chosen subset of DT leaves using a soft adjustment mechanism that balances fairness (R2) and utility (R3). This contrasts with approaches such as \mostlyai\ which apply adjustments across the entire probability distribution and all subgroups, potentially impacting data utility more broadly.
\begin{algorithm}[t]
\caption{Fair Leaf Resampling}
\label{algo:resampling}
\begin{algorithmic}[1] 
\Statex \textbf{Input} $Tree$, $\lambda$ (fairness/utility tradeoff), $thr_{disc}=0$
\vspace{0.2cm} 
\Statex\textbf{Output} Fair $Tree$ \Comment{Tree with fair leaf sampling probabilities $p_{\ell}$}
\vspace{-0.2cm} 
\Statex \hrulefill  
\Statex \textbf{i) Greedy leaf search}
\State $\mathcal{C} = \{\ell \in Tree \ | \ \Delta \text{disc}_\ell < 0\}$ \Comment{Leaf candidates for relabeling}
\State $\mathcal{L} = \{\}$
\While{$\text{new\_disc}(\mathcal{L}) > thr_{disc}$} 
    \State $\text{best}_{\ell} := argmax_{\ell \in \mathcal{C}\setminus\mathcal{L}}(\text{disc}_{\ell}/\text{acc}_{\ell})$
    \State $\mathcal{L} \gets \mathcal{L} \cup \text{best}_{\ell}$
\EndWhile
\vspace{-0.2cm} 
\Statex \hrulefill  
\Statex \textbf{ii) Resampling}
\For{$\ell \in \mathcal{L}$} 
    \State $p_{\ell} \gets  p_{\ell} * (1 - \lambda) + (1 - p_{\ell}) * \lambda$ \Comment{Set new sampling probabilities for leaf}
\EndFor
\State \textbf{Return} $Tree$
\end{algorithmic}
\end{algorithm}

\section{Experiments}

\subsection{Experimental Setup}
To ensure robustness, we perform 3-fold data splits, thus having $66\%$ of the data for training and $33\%$ for testing on unseen real data.
These splits are performed at the very beginning of the pipeline, meaning that, for each fold, all generative models are learned on $66\%$ of the whole data and that the real data test set for the current split remains unseen, even during the generation. We found that variability across different splits was much more significant than variability from repeating generative runs within a split, so we use one run per split and report average results with standard deviation across the three folds. Furthermore, splits are kept the same for all generative methods, ensuring that all models have been trained and tested on the same real data.

As stated before, \us\ uses the original order of features from the given dataset except when mentioned otherwise.
For downstream evaluation, we use LightGBM \cite{ke2017lightgbm}, a SOTA gradient boosting model for tabular data classification, though results remain consistent with other classifiers (not reported due to space constraints). Runtime experiments are conducted on a CPU for our approach, while methods supporting GPU acceleration are run on a GPU. Our hardware specifications include a 12th Gen Intel(R) Core(TM) i9 processor CPU and an Nvidia GeForce RTX 3080 Ti GPU.

\subsection{Datasets}
\label{sec:datasets}


\begin{table}[t]
\caption{Dataset characteristics.}
\label{tab:datasets}

\centering
\setlength{\tabcolsep}{4pt} 
\begin{tabular}{lcccc}
\midrule
\textbf{Dataset} & \makecell[c]{\textbf{Num.}\\ \textbf{Samples}} & \makecell[c]{\textbf{Num. Feat.}\\\textbf{(Num/Cat)}} & \makecell[c]{\textbf{Sensitive}\\ \textbf{Attribute} $\mathbf{S}$} & \makecell[c]{\textbf{Target Class}\\ $\mathbf{Y}$} \\ \hline
\makecell[l]{Adult\\Census} & 45k & 4/8 & 
\multicolumn{2}{c}{
    \begin{tabular}{cc}
    sex & income \\ \midrule
           & $<50K$ \quad $\geq50K$ \\
    \cline{2-2}
    female & 28.80\% / 3.69\% \\
    male   & 46.41\% / 21.09\% \\
    \end{tabular}
} \\ \hline

\makecell[l]{Dutch\\Census} & 60k & 0/11 & 
\multicolumn{2}{c}{
    \begin{tabular}{cc}
    sex & occupation level \\ \midrule
           & low \quad high \\
    \cline{2-2}
    female & 33.71\% / 16.39\% \\
    male   & 18.68\% / 31.21\% \\
    \end{tabular}
} \\ \hline

\makecell[l]{Bank\\Marketing} & 40k & 7/9 & 
\multicolumn{2}{c}{
    \begin{tabular}{cc}
    marital-st. & deposit \\ \midrule
           & no \quad yes \\
    \cline{2-2}
    married & 61.14\% / 6.88\% \\
    non-married   & 27.19\% / 4.77\% \\
    \end{tabular}
} \\ \hline

\makecell[l]{KDD\\Census} & 95k & 7/31 & 
\multicolumn{2}{c}{
    \begin{tabular}{cc}
    sex & income \\ \midrule
           & low \quad high \\
    \cline{2-2}
    female & 50.81\% / 1.21\% \\
    male   & 43.42\% / 4.54\% \\
    \end{tabular}
} \\ \hline

\makecell[l]{ACS-I\\Utah} & 19k & 2/7 & 
\multicolumn{2}{c}{
    \begin{tabular}{cc}
    sex & income \\ \midrule
           & low \quad high \\
    \cline{2-2}
    female & 32.5\% / 13.74\% \\
    male   & 23.78\% / 29.96\% \\
    \end{tabular}
} \\ \hline

\makecell[l]{ACS-I\\Alabama} & 24k & 2/7 & 
\multicolumn{2}{c}{
    \begin{tabular}{cc}
    sex & income \\ \midrule
           & low \quad high \\
    \cline{2-2}
    female & 34.08\% / 14.31\% \\
    male   & 26.17\% / 25.42\% \\
    \end{tabular}
} \\ \hline
\end{tabular}
\end{table}

For our experimental evaluation, we selected six publicly available tabular datasets based on relevant surveys \cite{han2023ffb,le2022survey} that identify datasets suitable for fairness assessment. We list the characteristics of all datasets in Table~\ref{tab:datasets}, such as the total number of samples, the number of numerical and categorical features, as well as the distribution of the four subgroups defined by the sensitive attribute, and the target.
It is important to mention that the Dutch Census dataset consists exclusively of categorical features. Additionally, the KDD Census dataset is the largest in our experiments, containing the highest number of samples. It is primarily composed of categorical features and exhibits a significant class imbalance, which makes it the most skewed dataset considered for the experiments. The last two datasets, namely ACS-I Utah and Alabama, are derived from the \emph{new Adult} dataset \cite{ding2021retiring}. These states were selected by ranking all states, with an adequate number of samples, based on the difference in positive outcome rates between sensitive groups.




\subsection{Baseline Competitors}
\label{sec:baseline_competitors}

We compare \us\ with the following state-of-the-art (SOTA) fair generative methods, along with one data augmentation technique. Like our approach, all these methods aim to optimize for downstream statistical parity. We do not compare against tabular data generators that solely optimize for utility, as they do not incorporate fairness objectives.

\noindent\textbf{\mostlyai}
\cite{tiwald2025tabularargn} is a recent, commercially developed, SOTA deep autoregressive method. It first discretizes all features before training the autoregressive model using a random order for the features of each batch. This allows conditional generation, where one or more features may be fixed to given values, and the model generates the remaining feature values. Like our approach, fairness is enforced only at the final step during target generation \cite{krchova2023strong}, by aligning the full conditional distributions across sensitive groups. In contrast, our method applies minimal and localized interventions, targeting only a subset of leaves selected greedily by our resampling algorithm.

\noindent{\textbf{\tabfairgan}}
\cite{rajabi2022tabfairgan} is a deep method based on GANs, incorporating fairness in the training process, which is done in two phases. First, the model undergoes normal training to learn the original data distribution, then the loss function is modified to incorporate fairness, optimizing for statistical parity.

\noindent{\textbf{\cuts}
\cite{vero2023cuts} is another deep approach developed to be a customizable generator. Training is performed in two phases: the model is pre-trained to generate data that closely matches the distribution of the real samples, and then a second, user-defined, objective is used to fine-tune the model. Various objectives can be used in the second phase, including fairness, which is relevant to our work, e.g., statistical parity.

\noindent{\textbf{\fsmote}
\cite{chakraborty2021bias} is a SMOTE-based \cite{chawla2002smote} data augmentation method that augments minority subgroups from the target and the sensitive attribute to address class and group imbalance.

\noindent{\textbf{\prefair}
\cite{pujol2022prefair} uses a deep architecture to generate private and fair data. It leverages differential privacy and causal fairness constraints to prevent unjust causal relationships between the sensitive attribute and the target. 

\subsection{Metrics}
\label{sec:metrics}

We evaluate the generated data by measuring (i) utility, i.e. the predictive performance for a downstream task (R3), (ii) fairness, in terms of statistical parity for the downstream task (R2), and (iii) data quality, in terms of distribution and similarity between real and fake samples (R1).

\noindent{\textbf{Utility:}} We train a classifier on the synthetic data $f(\hat{X}) \rightarrow \hat{Y}$, and evaluate on real unseen test data, measuring the ROC AUC score of the predictions $Y_{pred}$.

\noindent{\textbf{Fairness:}} We use the same trained classifier and real test data to evaluate for fairness, by comparing the conditional probabilities between subgroups of $S$ in the test set. Specifically, we measure statistical parity, defined as: 
\[\text{Stat. Par.} := P(Y_{pred}=1 | S=0) - P(Y_{pred}=1 | S=1)\]
Our goal is to minimize this difference, ensuring that the predicted positive outcome rates are the same across subgroups.


\noindent{\textbf{Data quality:}} First, we train a classifier on $D \cup \hat{D}$ to distinguish between real and synthetic samples. The quality of the generated data is then evaluated using the ROC AUC score, which serves as a detection metric. Ideally, we aim for a detection score (Det. Sc.) close to $0.5$, indicating that the classifier struggles to differentiate between real and synthetic samples. Furthermore, we study the overall quality of the generated data by comparing their statistical distribution with the original samples. We use the Kolmogorov-Smirnov (KS) score to evaluate this with continuous features, which computes the maximal difference between the cumulative density functions of the real and synthetic data, and the Total Variation (TV) score for categorical features, which compares the probabilities for all the categorical values of a feature. In addition, we consider precision, recall \cite{sajjadi2018assessing}, density, and coverage \cite{naeem2020reliable} scores. All four of these metrics use neighborhood spheres computed using the closest data point. Precision (Prec.) gets how many of the synthetic data fall in at least one real sphere, while recall (Rec.) computes the inverse, how many real data fall in at least one synthetic sphere. Density (Dens.) is a variation of precision to take outliers into account. It computes how many of the real spheres a synthetic point appears in. Finally, coverage (Cov.) is a variation of recall and calculates the fraction of the real neighborhood spheres that contain a synthetic data point.
As these last four metrics may be fooled by generated samples being duplicates of the real ones, we also consider the median Distance to Closest Record (DCR) \cite{danholt2024syntheval}.
This metric uses nearest neighbors to compute the ratio between median distances from generated samples to real neighboring samples and from the real samples to their own nearest neighbors.
DCR ranges from 0 and is unbounded from above, with values smaller than 1 indicating that the generated data are closer to the real ones, while larger values indicate generated data further away from the real samples.

\section{Results}


\begin{figure*}[t]
    \centering
    \includegraphics[width=0.78\textwidth]{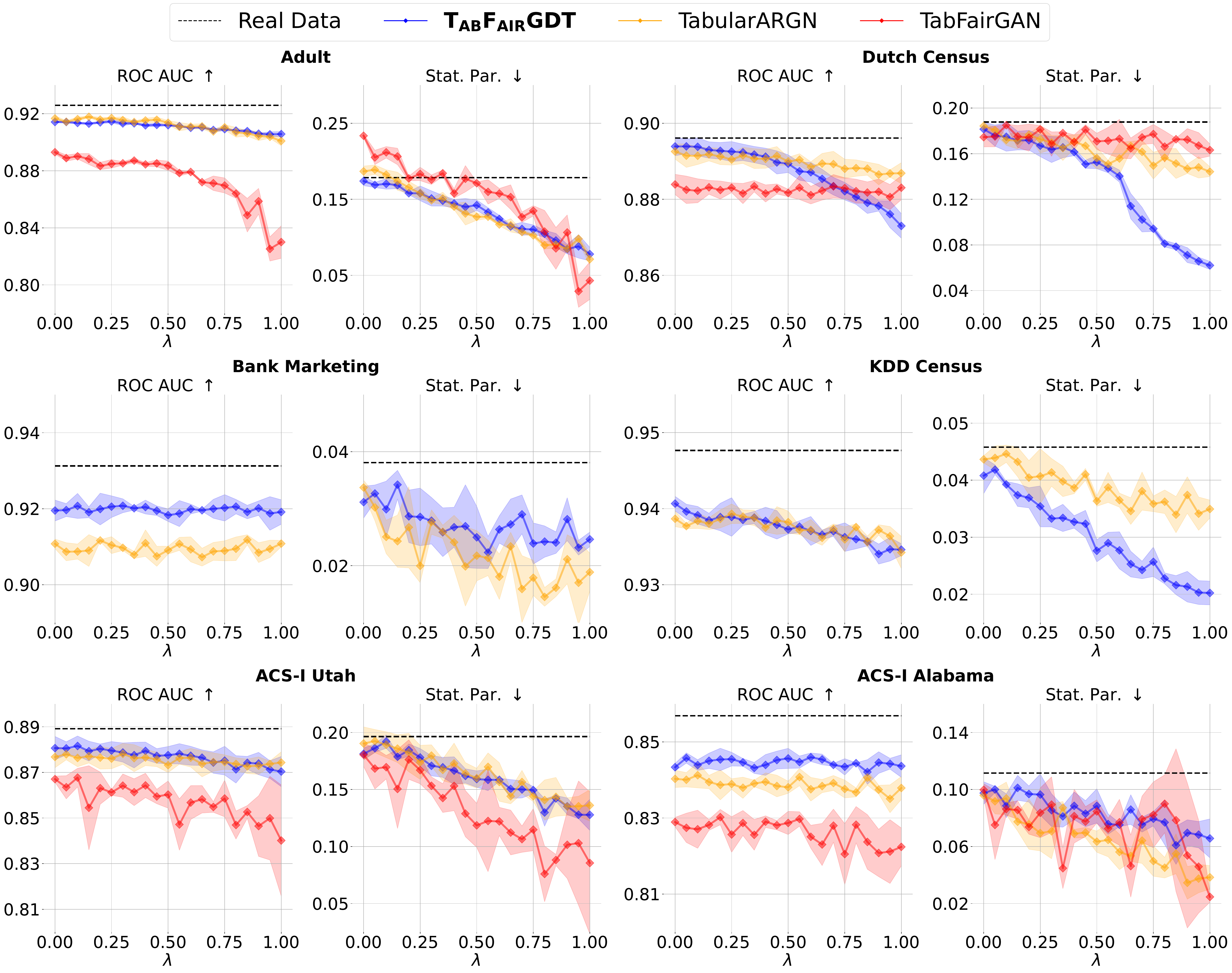}
    \caption{Impact of the fairness-utility tradeoff parameter $\lambda$ on model performance across all datasets. As $\lambda$ increases, models prioritize fairness, often at the expense of utility.}
    
    \label{fig:lamda_results}
\end{figure*}

We begin by examining the fairness-utility tradeoff (requirements R2-R3 described in Section~\ref{sec:problem_def}). Specifically, we compare our method, \us, to training on the real data, and to two competing deep models, \mostlyai\ and \tabfairgan, both of which also incorporate a fairness-utility tradeoff parameter $\lambda \in [0,1]$. Figure~\ref{fig:lamda_results} showcases how this tradeoff parameter affects both utility (ROC AUC) and fairness (Stat. Par.). All three methods perform mostly as anticipated, balancing utility retention with a reduction in discrimination as $\lambda$ increases. However, \tabfairgan\ fails for the KDD Census and Bank Marketing datasets, which are higher-dimensional, and highly class-imbalanced (see Section~\ref{sec:datasets}). In fact, it results in worse fairness compared to training on real data, as shown in Table~\ref{tab:all_datasets_utility_fairness}. Consequently, we exclude its plot for these datasets. We highlight that, even in this highly imbalanced case, our method can improve fairness with a minor loss in utility.
Additionally, for the Dutch census dataset, which consists solely of discrete features, \us\ is the only method that achieves a notable fairness improvement, while loosing only $2\%$ of utility. This further demonstrates the advantage of our method's non-parametric approach, which offers greater flexibility and can handle discrete features with no need for pre-processing. For the remainder of our results, we set the utility-tradeoff hyperparameter to $\lambda=1$ (for optimal fairness) for all methods.

\begin{table}[t]

\setlength{\tabcolsep}{12pt} 

\centering
\caption{Percentile difference of utility and fairness compared to training on real data, for all methods, averaged across all datasets. Methods that fail in at least one dataset are indicated in red.}

\begin{tabular}{l|cc}
\hline
 & \textbf{Utility} & \textbf{Fairness} \\
 \hline
\textbf{Method} & ROC AUC $\uparrow$ & Stat. Par. $\downarrow$ \\

\hline

\textbf{\us} & \textbf{-1.85\%} {\tiny $\pm$ 0.47\%} & \textbf{-48.41\%} {\tiny $\pm$ 12.03\%} \\
\mostlyai & -1.87\% {\tiny $\pm$ 0.56\%} & -42.28\% {\tiny $\pm$ 17.17\%} \\
\cdashline{1-3}\\[-1.5ex]
{\color{red}\tabfairgan} & -9.53\% {\tiny $\pm$ 7.15\%} & +26.47\% {\tiny $\pm$ 132.57\%} \\
{\color{red}\cuts} & -5.57\% {\tiny $\pm$ 4.55\%} & -44.59\% {\tiny $\pm$ 23.69\%} \\
{\color{red}\fsmote} & -0.50\% {\tiny $\pm$ 0.12\%} & -24.58\% {\tiny $\pm$ 5.38\%} \\
{\color{red}\prefair} & -15.71\% {\tiny $\pm$ 3.19\%} & -66.55\% {\tiny $\pm$ 14.40\%} \\
\hline
\end{tabular}
\label{tab:results_percentile}
\end{table}

\begin{table*}
\centering
\caption{Results for data quality metrics averaged across all datasets}
\label{tab:average_quality_metrics}
\begin{tabular}{l|cccccccc}
\hline
\textbf{Method} & \textbf{Det. Sc. $\approx 0.50$} & \textbf{KS $\uparrow$} & \textbf{TV $\uparrow$} & \textbf{Precision $\uparrow$} & \textbf{Recall $\uparrow$} & \textbf{Density $\uparrow$} & \textbf{Coverage $\uparrow$} & \textbf{DCR $\approx 1.00$} \\
\hline
\textbf{\us} & \textbf{0.54} {\tiny$\pm$0.0} & \textbf{0.83}{\tiny$\pm$0.4} & \textbf{0.99} {\tiny$\pm$0.0} & 0.82 {\tiny$\pm$0.2} & 0.82 {\tiny$\pm$0.2} & \textbf{0.79} {\tiny$\pm$0.3} & \textbf{0.82} {\tiny$\pm$0.2} & \textbf{1.02} {\tiny$\pm$0.0} \\
\mostlyai & 0.60 {\tiny$\pm$0.0} & 0.82 {\tiny$\pm$0.4} & 0.98 {\tiny$\pm$0.0} & 0.81 {\tiny$\pm$0.2} & \textbf{0.84} {\tiny$\pm$0.2} & 0.73 {\tiny$\pm$0.3} & 0.81 {\tiny$\pm$0.2} & 1.27 {\tiny$\pm$0.2} \\
\tabfairgan & 0.83 {\tiny$\pm$0.1} & 0.75 {\tiny$\pm$0.4} & 0.96 {\tiny$\pm$0.0} & 0.61 {\tiny$\pm$0.3} & 0.71 {\tiny$\pm$0.3} & 0.48 {\tiny$\pm$0.3} & 0.57 {\tiny$\pm$0.3} & 2.44 {\tiny$\pm$1.1} \\
\cuts & 1.00 {\tiny$\pm$0.0} & 0.49 {\tiny$\pm$0.3} & 0.62 {\tiny$\pm$0.4} & 0.28 {\tiny$\pm$0.4} & 0.31 {\tiny$\pm$0.5} & 0.24 {\tiny$\pm$0.4} & 0.27 {\tiny$\pm$0.4} & 18.21 {\tiny$\pm$25.0} \\
\fsmote & 0.64 {\tiny$\pm$0.1} & 0.77 {\tiny$\pm$0.4} & 0.95 {\tiny$\pm$0.0} & \textbf{0.83} {\tiny$\pm$0.2} & 0.82 {\tiny$\pm$0.2} & 0.76 {\tiny$\pm$0.3} & 0.79 {\tiny$\pm$0.2} & 1.08 {\tiny$\pm$0.2} \\
\prefair & 0.93 {\tiny$\pm$0.1} & 0.71 {\tiny$\pm$0.4} & 0.97 {\tiny$\pm$0.0} & 0.55 {\tiny$\pm$0.2} & 0.77 {\tiny$\pm$0.2} & 0.34 {\tiny$\pm$0.2} & 0.44 {\tiny$\pm$0.2} & 4.40 {\tiny$\pm$2.1} \\
\hline
\end{tabular}
\end{table*}

\begin{table}[t]
\setlength{\tabcolsep}{3pt} 
\centering
\caption{Results per dataset for all methods. Failures are indicated in red color.}
\begin{tabular}{l|cccc}

\hline
\textbf{Method} & ROC AUC $\uparrow$ & Stat. Par. $\downarrow$ & ROC AUC $\uparrow$ & Stat. Par. $\downarrow$ \\
\hline
 & \multicolumn{2}{c}{Adult} & \multicolumn{2}{c}{Dutch Census} \\

\hline
Real Data & 0.926 {\tiny $\pm$ 0.001} & 0.178 {\tiny $\pm$ 0.007} & 0.896 {\tiny $\pm$ 0.003} & 0.188 {\tiny $\pm$ 0.005} \\
\cdashline{1-5}\\[-2ex] 
\textbf{\us} & 0.906 {\tiny $\pm$ 0.002} & 0.078 {\tiny $\pm$ 0.009} & 0.873 {\tiny $\pm$ 0.003} & 0.062 {\tiny $\pm$ 0.004} \\
\mostlyai & 0.901 {\tiny $\pm$ 0.003} & 0.071 {\tiny $\pm$ 0.007} & 0.887 {\tiny $\pm$ 0.003} & 0.144 {\tiny $\pm$ 0.004} \\
\tabfairgan & 0.830 {\tiny $\pm$ 0.011} & 0.043 {\tiny $\pm$ 0.025} & 0.883 {\tiny $\pm$ 0.003} & 0.163 {\tiny $\pm$ 0.005} \\
\cuts & 0.905 {\tiny $\pm$ 0.003} & 0.145 {\tiny $\pm$ 0.006} & {\color{red}0.500} {\tiny $\pm$ 0.000} & - \\
\fsmote & 0.919 {\tiny $\pm$ 0.001} & {\color{red}0.225} {\tiny $\pm$ 0.003} & 0.893 {\tiny $\pm$ 0.003} & 0.153 {\tiny $\pm$ 0.008} \\
\prefair & 0.760 {\tiny $\pm$ 0.012} & 0.057 {\tiny $\pm$ 0.012} & 0.804 {\tiny $\pm$ 0.003} & 0.105 {\tiny $\pm$ 0.001} \\
\hline
 & \multicolumn{2}{c}{Bank Marketing} & \multicolumn{2}{c}{KDD Census} \\
\hline

Real Data & 0.931 {\tiny $\pm$ 0.001} & 0.038 {\tiny $\pm$ 0.003} & 0.948 {\tiny $\pm$ 0.001} & 0.046 {\tiny $\pm$ 0.002} \\
\cdashline{1-5}\\[-2ex]  
\textbf{\us} & 0.919 {\tiny $\pm$ 0.003} & 0.025 {\tiny $\pm$ 0.001} & 0.935 {\tiny $\pm$ 0.001} & 0.020 {\tiny $\pm$ 0.002} \\
\mostlyai & 0.911 {\tiny $\pm$ 0.001} & 0.019 {\tiny $\pm$ 0.003} & 0.934 {\tiny $\pm$ 0.002} & 0.035 {\tiny $\pm$ 0.002} \\
\tabfairgan & 0.716 {\tiny $\pm$ 0.012} & {\color{red}0.071} {\tiny $\pm$ 0.017} & 0.827 {\tiny $\pm$ 0.028} & {\color{red}0.181} {\tiny $\pm$ 0.036} \\
\cuts & 0.896 {\tiny $\pm$ 0.001} & 0.016 {\tiny $\pm$ 0.004} & 0.925 {\tiny $\pm$ 0.002} & 0.008 {\tiny $\pm$ 0.001} \\
\fsmote & 0.919 {\tiny $\pm$ 0.003} & {\color{red}0.051} {\tiny $\pm$ 0.003} & 0.937 {\tiny $\pm$ 0.001} & {\color{red}0.060} {\tiny $\pm$ 0.001} \\
\prefair & {\color{red}0.511} {\tiny $\pm$ 0.017} & - & {\color{red}0.613} {\tiny $\pm$ 0.118} & - \\
\hline

 & \multicolumn{2}{c}{ACS-I Utah} & \multicolumn{2}{c}{ACS-I Alabama} \\

\hline
Real Data & 0.889 {\tiny $\pm$ 0.004} & 0.196 {\tiny $\pm$ 0.008} & 0.857 {\tiny $\pm$ 0.001} & 0.111 {\tiny $\pm$ 0.005} \\
\cdashline{1-5}\\[-2ex]  
\textbf{\us} & 0.870 {\tiny $\pm$ 0.007} & 0.128 {\tiny $\pm$ 0.014} & 0.844 {\tiny $\pm$ 0.003} & 0.066 {\tiny $\pm$ 0.014} \\
\mostlyai & 0.874 {\tiny $\pm$ 0.005} & 0.136 {\tiny $\pm$ 0.014} & 0.838 {\tiny $\pm$ 0.003} & 0.038 {\tiny $\pm$ 0.008} \\
\tabfairgan & 0.840 {\tiny $\pm$ 0.024} & 0.086 {\tiny $\pm$ 0.062} & 0.822 {\tiny $\pm$ 0.005} & 0.025 {\tiny $\pm$ 0.004} \\
\cuts & 0.845 {\tiny $\pm$ 0.008} & 0.148 {\tiny $\pm$ 0.043} & 0.733 {\tiny $\pm$ 0.029} & 0.069 {\tiny $\pm$ 0.052} \\
\fsmote & 0.885 {\tiny $\pm$ 0.005} & 0.150 {\tiny $\pm$ 0.010} & 0.851 {\tiny $\pm$ 0.001} & 0.076 {\tiny $\pm$ 0.010} \\
\prefair & 0.729 {\tiny $\pm$ 0.041} & 0.032 {\tiny $\pm$ 0.029} & 0.715 {\tiny $\pm$ 0.029} & 0.033 {\tiny $\pm$ 0.011} \\
\hline
\end{tabular}
\label{tab:all_datasets_utility_fairness}
\end{table}



\begin{table}[t]
\caption{Computational runtime for fitting, sampling, and total time (in seconds)}
\label{tab:generation_times}
\begin{subtable}{\columnwidth}
\centering
\caption{For all methods on  a dataset of 10 features and 10k samples}
\label{tab:generation_times_all}
\begin{tabular}{l|ccc}
\hline
\textbf{Method} & \textbf{Fitting} & \textbf{Sampling} & \textbf{Total} \\
\hline
\textbf{\us} & 0.81 & 0.25 & \textbf{1.05} \\
\mostlyai & 10.72 & 0.12 & 10.84 \\
\tabfairgan & 71.74 & \textbf{0.008} & 71.75 \\
\cuts & 882.44 & 0.20 & 882.65 \\
\fsmote & 0.018 & 29.17 & 29.19 \\
\prefair & \textbf{0.005} & 45.49 & 45.5 \\
\hline
\end{tabular}
\end{subtable}

\vspace{10pt}

\begin{subtable}{\columnwidth}
\setlength{\tabcolsep}{1pt} 
\centering
\caption{For \us\ vs \mostlyai\ on varying dataset sizes}
\label{tab:generation_times_varying_sizes}
\begin{tabular}{llc|ccc|ccc}
\toprule
\multicolumn{3}{l|}{\textbf{Dataset size}} & \multicolumn{3}{c|}{\textbf{\us}} & \multicolumn{3}{c}{\mostlyai} \\
\hline
\textbf{Num. Feat.} & \textbf{Num. Samp.} & & \textbf{Fit.} & \textbf{Samp.} & \textbf{Tot.} & \textbf{Fit.} & \textbf{Samp.} & \textbf{Tot.} \\ \midrule
\multirow{3}{*}{$10$} & 1k & & \textbf{0.15} & \textbf{0.06} & \textbf{0.21} & 10.88 & 0.09 & 10.97  \\
 & 10k & & \textbf{0.81} & 0.25 & \textbf{1.05} & 10.72 & \textbf{0.12} & 10.84  \\
 & 50k & & \textbf{3.84} & 1.04 & \textbf{4.88} & 27.18 & \textbf{0.33} & 27.51  \\
\midrule
\multirow{3}{*}{$100$} & 1k & & \textbf{2.76} & 1.03 & \textbf{3.79} & 36.75 & \textbf{0.75} & 37.50  \\
 & 10k & & \textbf{14.27} & 3.70 & \textbf{17.97} & 36.37 & \textbf{1.15} & 37.52  \\
 & 50k & & \textbf{70.11} & 15.88 & \textbf{85.99} & 126.71 & \textbf{3.25} & 129.96  \\
\midrule
\multirow{3}{*}{$500$} & 1k & & \textbf{33.43} & 18.48 & \textbf{51.91} & 206.90 & \textbf{9.38} & 216.28  \\
 & 10k & & \textbf{167.08} & 43.29 & \textbf{210.37} & 580.68 & \textbf{13.78} & 594.46  \\
 & 50k & & \textbf{897.80} & 174.12 & \textbf{1071.92} & 2936.21 & \textbf{31.74} & 2967.95  \\
\midrule
\multicolumn{3}{l|}{\makecell[l]{Average speedup (\%)}} & \textbf{{\tiny+}77}{\tiny$\pm$16} & {\small-}188{\tiny$\pm$147} & \textbf{{\tiny+}72}{\tiny$\pm$19} & & &  \\
\bottomrule
\end{tabular}
\end{subtable}

\end{table}

Table~\ref{tab:results_percentile} provides a summary of our results, averaged across all datasets, in terms of percentile utility loss and fairness improvement, compared to training on real data. The more detailed per-dataset results are presented in Table~\ref{tab:all_datasets_utility_fairness}. To ensure a fair comparison, the aggregated results of Table~\ref{tab:results_percentile} exclude failed cases, i.e. where a method results in a utility loss greater than $30\%$ or fails to improve statistical parity relative to real data. We highlight in red the methods that fail on at least one dataset. From the results, we observe that our method, \us, leads to a major improvement in fairness, reducing statistical parity by almost $50\%$ on average, while losing less than $2\%$ in utility. Moreover, \us\ is more consistent across datasets, with a standard deviation of only $12\%$ for fairness, compared to much higher variability observed in other methods. \tabfairgan, in particular, shows exceptionally high standard deviation, indicating inconsistent performance across datasets. This suggests its effectiveness is highly sensitive to dataset characteristics, limiting practical reliability. Furthermore, our method, together with \mostlyai, are the only ones that succeed in producing realistic and fair synthetic data for all datasets.
Although \prefair\ performs good for fairness, it also results in large drops in utility. 
Notably, as observed in Table~\ref{tab:all_datasets_utility_fairness}, our method is the only one that succeeds in improving fairness for the Dutch Census dataset, which has only categorical features. We attribute this success to our non-parametric modeling approach.

We continue our evaluation by focusing on generated data quality (requirement R1). As reported in Table~\ref{tab:average_quality_metrics}, our method is the clear winner, achieving a detection score of only $0.54\%$.
For the other metrics, \us\ demonstrates optimal scores and generally surpasses the other deep models. Furthermore, the average DCR score of $1.02$ indicates that the synthetic data are as close to real samples as real samples are to each other, showing no signs of overfitting (which would correspond to a DCR below 1). 
We attribute these strong data quality results to the use of DTs, which effectively capture the structure of the training data, enabling the generation of synthetic data that closely aligns with the original distribution. 
Additionally, our method requires no pre-processing, unlike \mostlyai, which relies on discretization and rare-feature handling that can impact quality on certain datasets.

The results of our method can be summarized by considering the three requirements of fair synthetic data, listed in Section \ref{sec:problem_def}. First, \us\ can generate data that closely match the original samples, as shown by the metrics for data quality in Table \ref{tab:average_quality_metrics} (\textbf{R1}). Next, our method shows strong performance in reducing statistical parity for all datasets (Table \ref{tab:results_percentile} and Table \ref{tab:all_datasets_utility_fairness}), progressing towards statistical independence between the target and the sensitive attribute (\textbf{R2}). Finally, \us\ achieves this performance while having a minor impact on utility scores, as shown in Table \ref{tab:results_percentile} (\textbf{R3}).
Our method thus meets all three requirements needed to obtain fair synthetic data while using simple DTs. Moreover, it performs on average better and is more efficient than deep approaches that also meet requirements, such as \mostlyai, without the need for specialized hardware.

\begin{figure*}[t]
    \centering
    \includegraphics[width=0.72\textwidth]{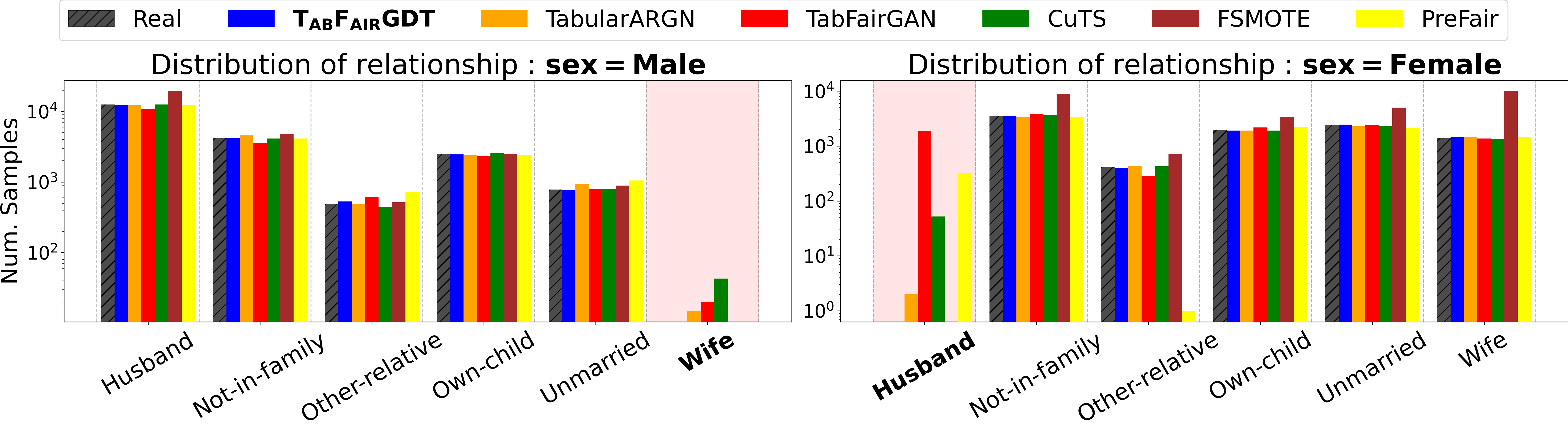}
    \caption{Per-group feature distributions of real vs. synthetic data for the \emph{relationship} attribute of the Adult census dataset. Areas with synthetic OOD samples are highlighted in red. Our method \us\ does not produce any OOD samples.}
    \label{fig:relationship_distribution}
\end{figure*}

\begin{figure*}[t]
    \centering
    \includegraphics[width=0.71\textwidth]{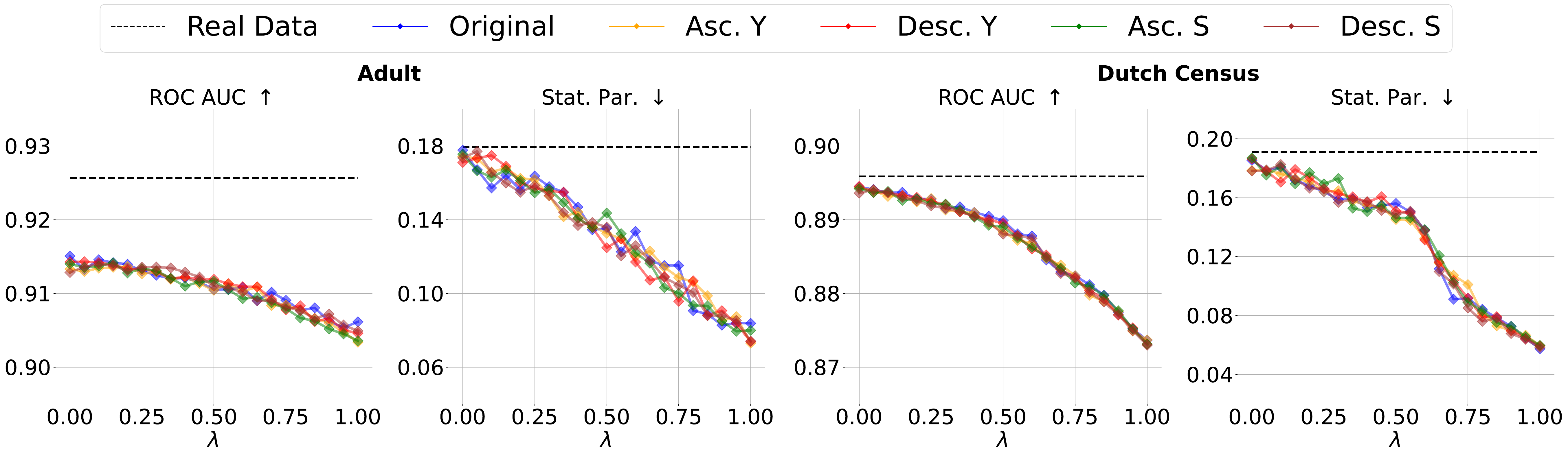}
    \caption{Effect of column ordering in \us\ on the utility-fairness tradeoff. Standard deviations were removed for clarity.}
    \label{fig:lamda_results_ordering}
\end{figure*}

\subsection{Computational Runtimes}

Our method is not only effective but also highly efficient and lightweight, operating entirely on the CPU. We evaluate all approaches on a medium-sized synthetic dataset with 10 features and 10k instances and report fitting, sampling, and total runtime in Table~\ref{tab:generation_times_all}. To ensure a fair comparison, we utilize GPU acceleration for methods that support it. Our results show that \us\ achieves the fastest total time, generating synthetic data in one second, followed by \mostlyai. Given that \mostlyai\ is a commercially developed product where efficiency is a priority, it incorporates an early stopping criterion during the fitting phase. Consequently, we select \mostlyai\ as the primary benchmark for further comparisons across varying dataset sizes, increasing both the number of features and instances, presented in Table~\ref{tab:generation_times_varying_sizes}. Once again, our method consistently demonstrates the lowest total time, outperforming \mostlyai\ by an average of $72\%$. This advantage is primarily due to our highly efficient fitting process, which enables parallel tree construction. Notably, fast fitting can be particularly valuable in real-world scenarios such as continuous model retraining during deployment, where low-latency updates are critical. However, as the dataset size increases, our method experiences slower sampling times due to the inherently sequential nature of the generation process.





\subsection{Generation of Out-Of-Distribution Samples}
\label{sec:female_husbands}

We analyze data from all methods to detect sensitive attribute-specific out-of-distribution (OOD) samples. Using the Adult census dataset as an example, we examine the \emph{relationship} feature by comparing real and synthetic distributions, per-group, in Figure~\ref{fig:relationship_distribution}. Notably, real data does \emph{not} contain instances where \{sex=Male, rel.=Wife\} or \{sex=Female, rel.=Husband\}. However, all baseline generative models introduce such OOD synthetic samples. Specifically, GAN-based methods \tabfairgan, \cuts, and \prefair, which first train a deep generative model and then fine-tune for fairness, generate a significant number of these instances. \mostlyai, despite using autoregressive modeling with fairness constraints applied only at the final step, also produces such samples, likely due to its random feature reordering. In contrast, our approach prevents this by generating $\hat{S}$ only after $\hat{X}$ is fully synthesized, and enforces fairness only for the target generation, not for features $X_j$. This prevents sampling from a distribution where $X_j \perp S$, thus respecting real-world constraints and avoiding unrealistic feature combinations. Interestingly, \fsmote\ also avoids this problem by generating synthetic samples via neighborhood-based interpolation.

Such OOD feature combinations might improve fairness by confusing downstream classifiers, but can be problematic in critical domains like healthcare. We argue that in scenarios involving subgroup-specific medical conditions (e.g., pregnancy, testicular cancer, etc.), it is crucial to ensure fairness without introducing OOD samples, especially when making the synthetic data publicly available.

\subsection{Impact of Feature Ordering in \us}
\label{sec:ordering}

Due to its autoregressive architecture at both fitting and generating times, one could argue that the order in which the features are considered may impact the results obtained with \us. The experiment in this subsection analyzes this factor by examining five possible orders: 1. the original order of the features in the dataset, 2, 3. ascending/descending order based on the correlation between the features and the target (Asc./Desc. $Y$), and 4, 5. ascending/descending order based on the correlation between the features and the protected attribute (Asc./Desc. $S$). Figure \ref{fig:lamda_results_ordering} shows that all feature orders produce nearly identical results, with no statistically significant differences observed. This trend was consistent across other datasets, and no meaningful changes were found in data quality metrics, so they are omitted here. We conclude that feature order does not affect the results obtained with \us.



\section{Conclusion and Future Work}

We introduced \us, a method for generating fair synthetic data using autoregressive DTs. \us\ produces realistic data, mitigates discrimination in downstream tasks, and maintains strong predictive performance.
Our method combines the simplicity and efficiency of DTs, outperforming deep generative models in terms of data quality, utility, and fairness. Furthermore, \us\ is up to $72\%$ faster on average, generating fair data for medium-sized datasets in just one second on a standard CPU. 

Future work includes extending our framework to support additional fairness definitions, intersectional fairness, regression tasks, and continuous sensitive attributes, as well as exploring hyperparameters, such as tree depth, to examine how over/under-fitting influences fairness and data quality. We also plan to use dataset similarity measures, e.g., the DT-partition-based approach in \cite{ntoutsi2008general}, to assess synthetic data quality, and the impact of fairness thresholds on the generated distributions.



\section*{Acknowledgment} This work is funded by the EU Horizon Europe project MAMMOth under contract number 101070285. We further acknowledge the support by the EU Horizon
Europe Project STELAR, Grant Agreement ID: 101070122.
Computational resources have been provided by the supercomputing facilities of the UCLouvain (CISM/UCL) and the Consortium des Équipements de Calcul Intensif en Fédération Wallonie Bruxelles (CÉCI) funded by the Fond de la Recherche Scientifique de Belgique (F.R.S.-FNRS) under convention 2.5020.11 and by the Walloon Region.


\bibliographystyle{IEEEtran}
\bibliography{bibliography}

\end{document}